\documentclass[journal]{IEEEtran}

\usepackage{graphicx}
\usepackage{amsmath, amssymb}
\usepackage{multirow}
\usepackage{makecell}
\usepackage{latexsym}
\usepackage{pifont}
\usepackage{booktabs}
\usepackage{colortbl}
\usepackage{xcolor}
\usepackage[driverfallback=dvipdfm,
CJKbookmarks=true, bookmarksnumbered=true, bookmarksopen=true,
colorlinks=true,
pdfborder=001,
citecolor=blue, linkcolor=blue, anchorcolor=red, urlcolor=black
]{hyperref}

\newcommand{\xmark}{\ding{55}}
\newcommand{\cmark}{\ding{51}}

\definecolor{demphcolor1}{gray}{.6}
\newcommand{\demphs}[1]{\textcolor{demphcolor1}{#1}}

\definecolor{lightred}{RGB}{241,140,142}
\definecolor{lightblue}{RGB}{122,46,246}
\definecolor{sherrybrown}{RGB}{227,207,87}
\definecolor{sherrytan}{RGB}{210,180,140}

\newcommand{\lemphs}[1]{\textcolor{lightblue}{#1}}

\newcommand{\dtplus}[1]{\fontsize{7pt}{0.1em}\selectfont (\textbf{\textcolor{green!80}{#1}})}

\begin{document}
\title{Adapt and Align to Improve Zero-Shot Sketch-Based Image Retrieval}

\author{Shiyin~Dong,
        Mingrui~Zhu,
        Nannan~Wang,~\IEEEmembership{Member,~IEEE,}
        and~Xinbo~Gao,~\IEEEmembership{Senior Member,~IEEE}
        }



\maketitle

\begin{abstract}

    Zero-shot sketch-based image retrieval (ZS-SBIR) is challenging due to the cross-domain nature of sketches and photos, as well as the semantic gap between seen and unseen image distributions. Previous methods fine-tune the pre-trained models with various side information and learning strategies to learn a compact feature space that (\romannumeral1) is shared between the sketch and photo domains and (\romannumeral2) bridges seen and unseen classes. However, these efforts are inadequate in adapting domains and transferring knowledge from seen to unseen classes. In this paper, we present an effective \emph{``Adapt and Align''} approach to address the key challenges. Specifically, we insert implement-friendly and lightweight domain adapters to learn new abstract concepts of the sketch domain and improve cross-domain representation capabilities, which helps alleviate domain heterogeneity and balance the pre-training prior bias. Remarkably, when only fine-tuning these adapters, we achieve higher mAP than previous the best full fine-tuned model (69.0 vs 68.8). Secondly, inspired by recent advances in image-text foundation models (\textit{e.g.}, CLIP) on zero-shot scenarios, we explicitly align the learned image embedding with a more semantic text embedding to fill semantic gap and achieve the desired knowledge transfer from seen to unseen classes. We successfully demonstrate the effectiveness of the proposed method on two widely-used model architectures (CNN and ViT). Extensive experiments on three benchmark datasets demonstrate the superiority of our method in terms of retrieval accuracy and flexibility.
  
\end{abstract}

\IEEEpeerreviewmaketitle

\section{Introduction}
    
    Freehand sketches can represent abstract semantic concepts with simple strokes. With the widespread adoption of touchscreen mobile devices, sketch-based image retrieval (SBIR) now has convenient application scenarios and significant value in multimedia community.  Formally, SBIR involves retrieving photos from an extensive gallery of images that belong to the same class as the given query sketch. However, sketches and photos come from different domains\footnote{Modality is a larger concept than domain in general definition. So, we use domain to describe photo and sketch and modality to describe image and language.} and may exhibit significant differences in the feature space even when they share the same class. Recently, only considering overcoming domain heterogeneity and assuming that all test classes are visible during training, several methods~\cite{LiLiu2017DeepSH, sangkloy2016sketchy, sain2021stylemeup, bhunia2021more,lu2018learning} have shown promising retrieval results. However, a more realistic and attractive scenario is that the test set categories are not visible during training, which is defined as zero-shot sketch-based image retrieval (ZS-SBIR).
    
    \begin{figure}[ht]
        \centering
        \includegraphics[width=1.00 \linewidth]{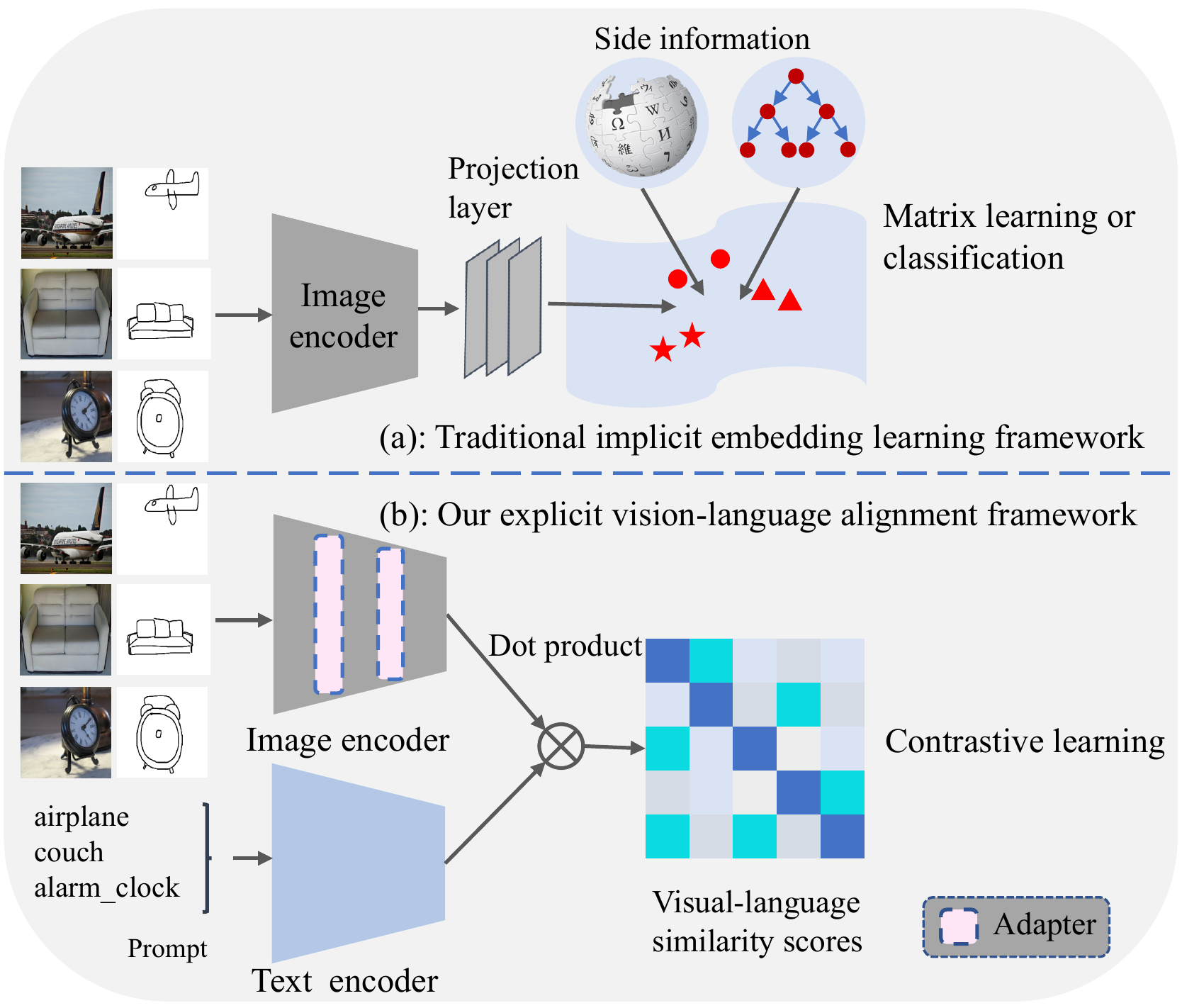}
        \caption{An illustration of (a) the existing pipeline and (b) our proposed method. Conventional techniques fine-tune pre-trained models and implicitly map semantic vectors to a common features space. In contrast, our pipeline employs adapters to bridge the domain gap and explicitly aligns the learned image embedding with a more semantic text embedding.}
        \label{fig: sherry_teaser}
    \end{figure}
    
    ZS-SBIR faces the same challenges as SBIR due to the domain gap between sketches and photos. However, it also presents an additional challenge due to its stringent definition, which involves transferring knowledge learned from seen classes to unseen classes. Previous solutions address the above challenges with different strategies. The first~\cite{deng2020progressive, zhan2022three-stream} projects samples from different domains into a shared space to minimize domain differences. The second~\cite {SounakDey2019DoodleTS, wang2021transferable} incorporates side information to embed semantic knowledge into visual features to learn knowledge transformation from seen to unseen classes. However, their approaches are still traditional and inefficient as they only focused on learning pure image information. As illustrated in Figure~\ref{fig: sherry_teaser} (a), they framed the training of the ZS-SBIR model as either a classification~\cite{liu2019semantic, tursun2022efficient, wang2022prototype} problem that mapped features to one-hot vectors or metric learning~\cite{SounakDey2019DoodleTS, wang2021transferable, dutta2020styleguide} problem that used negative sample mining to learn samples similarity. 
    
    Firstly, to overcome the drawback of inefficient training, a research direction termed parameter-efficient tuning has been trending in natural language processing (NLP)~\cite{adapter, hu2021lora}. The goal is only to insert and fine-tune some lightweight layers to keep generalization ability and effectively adapt to downstream tasks. We believe this idea is helpful for ZS-SBIR task. Secondly, although researchers have extensively studied image features, obtaining semantic representations of natural images by image model remains challenging. In contrast, as the cornerstone of human civilization, language itself possesses highly semantic information. Therefore, a novel idea is to learn aligned image-text representations leveraging semantically rich text, which has been adopted in large visual-language models (\textit{e.g.}, CLIP~\cite{radford2021learning} and ALIGN~\cite{ALIGN}). Following this idea, we formulate the training of our model as an image-text matching problem to incorporate richer semantic information. The join method can be seen in Figure~\ref{fig: sherry_teaser} (b).

    Concretely, we propose an effective ``Adapt and Align'' approach that comprises two novel modules: effective adapters and vision-language alignment. Firstly, we insert a few learnable Adapter layers, enabling them to learn new abstract concepts of sketches, balance the domain gap, and improve cross-domain representation capabilities. Secondly, we explicitly align the learned image embedding with the more semantic text embedding extracted by CLIP, allowing us to transfer knowledge to unseen classes more efficiently. Specifically, we compute the dot product to determine the similarity score between vision and language features and achieve the feature alignment by contrastive learning. We conduct extensive experiments on two prominent model architectures (ResNet and ViT) and the results demonstrate the broad applicability of our method. For easy reference, we denote our model as \textbf{\textcolor{sherrybrown}{Sherry}}: A \textbf{\textcolor{sherrybrown}{S}}imple met\textbf{\textcolor{sherrybrown}{h}}od for ZS-SBIR using \textbf{\textcolor{sherrybrown}{e}}ffective adapte\textbf{\textcolor{sherrybrown}{r}}s and vision-language alignment st\textbf{\textcolor{sherrybrown}{r}}ateg\textbf{\textcolor{sherrybrown}{y}}. Our contributions can be summarized as follows:

    \begin{itemize}
        \item We propose effective domain adapters that address the generalizability problem of ZS-SBIR by focusing on better adaptation to new tasks. Our approach is broadly applicable across various pre-trained image models and straightforward to implement. 
        \item We demonstrate that directly aligning the image-text embedding can help transfer knowledge from seen to unseen classes. This simple yet effective strategy can leverage rich semantic information of large image-text foundation models.
        \item We conduct extensive experiments on three popular datasets and achieve state-of-the-art performance. Our key ideas are simple and generic; thus, they can exploit increasingly powerful foundation models going forward.
    \end{itemize}
    We hope to provide some inspiration on how to utilize the foundation model in zero-shot setting.
    
\section{RELATED WORK}

    \begin{figure*}[t]
        \centering
        \includegraphics[width= \textwidth]{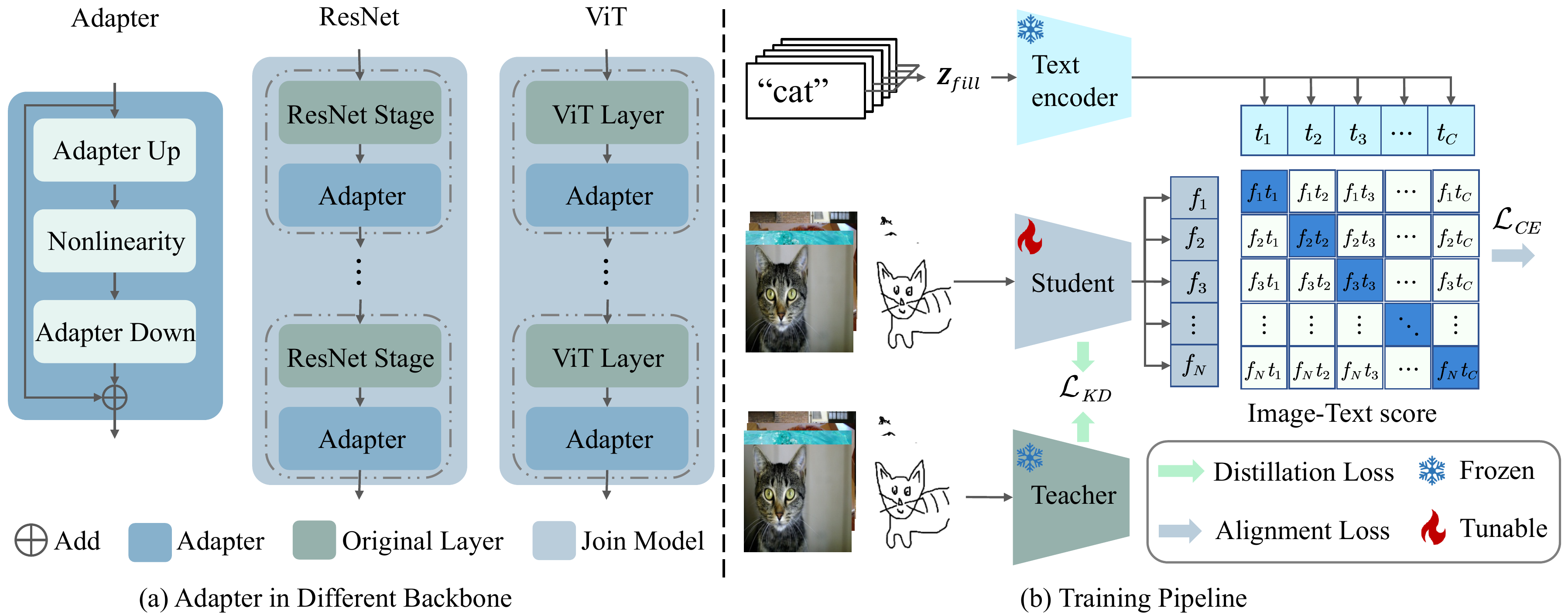}
        \caption{An overview of our method. On the left, we show how we insert adapters into standard model block. We insert an adapter after the original model block to learn efficient transfer and domian adaption. On the right is our alignment strategy. We first fill the template with category names and use a text encoder to extract text features. Then we align the image $f_k$ and text features $t_k$ by simple dot product and contrastive learning to achieve semantic transfer from seen to unseen classes.}
        \label{fig: framework for our method}
    \end{figure*}

\subsection{Zero-Shot Learning}

    The existing Zero-Shot Learning (ZSL) methods can be divided into two categories: embedding-based method~\cite{zhang2015zero} and generation-based method~\cite{verma2018generalized}.
    For embedding-based arts, most researchers project feature vectors to meaningful attribute vector designed by human~\cite{lampert2013attribute, akata2013label,lampert2009learning, ZSL-semantic, ZSL-Transferable-Contrastive} for the knowledge transfer from seen to unseen classes in the attribute space. 
    
    For the generation-based methods, most researchers focus on exploring new representation knowledge by learning latent distribution. For example, \cite{conditional-generative,feature-generating, ZSL-adversarial-embedding, ZSL-F-VAEGAN-D2, ZSL-Adaptive} used data augment methods to generate new data to mitigate the imbalance bias between seen and unseen samples. 

\subsection{Zero-Shot Sketch-Based Image Retrieval}

    Compared with SBIR~\cite{lu2018learning, sain2021stylemeup, bhunia2021more}, ZS-SBIR is a more challenging research topic, with two challenges: domain gap and semantic knowledge transfer. To mitigate problems of undesirable retrieval caused by domain heterogeneity, some researchers~\cite{deng2020progressive,tian2021relationship} believed that the circular consistency constraints across domains could mitigate domain differences. Dey \textit{et al}.~\cite{SounakDey2019DoodleTS} trained the classification network by inserting \emph{gradient reversal layer} (GRL)~\cite{ganin2015grl} to learn the domain-indistinguishable features. In addition, some methods~\cite{tursun2022efficient,jing2022augmented, dutta2020styleguide} learn cross-domain representation through triplet or quadruplet constraint to promote intra-class coherence and inter-class separability.

    To alleviate the knowledge transfer challenges in the Zero-Shot learning setting, some methods have utilized word embeddings\cite{mikolov2013word2vec, SounakDey2019DoodleTS} or hierarchical models\cite{miller1995wordnet}to migrate knowledge from seen to unseen classes~\cite{liu2019semantic, wang2021domain, wang2021transferable} or both of them~\cite{dutta2019sem-pcyc, deng2020progressive}. However, these methods used side information learned solely from the text corpus, neglecting  the interaction of image and language. 

    Liu \textit{et al}.~\cite{liu2019semantic} proposed a distillation learning strategy to overcome catastrophic forgetting and improve transferability. Based on this, Tian \textit{et al}.\cite{tian2021relationship} proposed learning local and contrastive relationships of the teacher model to explore distillation learning. Recently, some ViT~\cite{dosovitskiy2020ViT} based approachs~\cite{tian2022tvt, wang2022prototype} learned better global representation to improve generation ability. The most similar to our idea is CLIP for ALL Things~\cite{sain2023clip}, which made a meaningful attempt to use CLIP in the sketch-photo multimedia community. But it directly used CLIP\footnote{\href{https://github.com/openai/CLIP}{https://github.com/openai/CLIP}} ViT-B/32 model (pre-trained on a web-level dataset that contains 400 million image-text pairs) to retrieve, thus may result in potential label-leakage. We strictly follow the zero-shot setting as we use the model pre-trined on ImageNet-1k~\cite{ImageNet-1k} and make further exploration into how to make the lager vision language model beneficial in zero-shot setting.

\subsection{Effective Adaption}

    In natural language processing, Adapter~\cite{adapter} implemented a compact model that freezes the original network and adapted downstream tasks by adding only a few parameters, performing well in 26 language classification tasks. Several methods~\cite{MAD-X, UDP} demonstrated that the adapter could avoid forgetting past knowledge in a continual learning scene and capture the transferable features for the target domain while reserving the source domain knowledge. Their goal is to reduce the number of trainable parameters, thus lowering the computation cost while reaching or surpassing the fully fine-tuned model.

    In computer vision, DeiT~\cite{DA-DeiT} adds a novel distillation token to achieve time and date-efficient training and excels in downstream tasks. AIM~\cite{AIM} inserted different types of adapters for spatial and temporal adaption and achieved higher performance than previous fully fine-tuned huge video models. However, AIM kept the backbone frozen, but we made it tunable for better suitability for our task.
    
\subsection{Language Driven Model}

    Recently, CLIP (Contrastive Language Image Pre-training)~\cite{radford2021learning} learned high-level image and language representation by contrast learning on 400 million raw image-text pairs (called WIT) crawled from the Web and showed impressive transferability on 30 computer vision datasets. This model showed encouraging performance in zero-shot and few-shot settings in many 2D computer vision tasks~\cite{gao2021clip-adapter, rao2022denseclip, xu2022groupvit, gu2021ViLD} and other multi-modality tasks, \textit{e.g.}, video, speech, text~\cite{luo2022clip4clip, zhang2022pointclip, guzhov2022audioclip}. 
    
    CoOp~\cite{zhou2022CoOp} leveraged learnable textual tokens to acquire visual representations, demonstrating a more robust generalization ability in many vision tasks. LSeg~\cite{li2022L-seg} compute pixel-level visual-text alignment through the task-specific backbone and then predicted pixel labels and showed a desired performance in semantic segmentation tasks. ActionCLIP~\cite{wang2021actionclip} enhanced the traditional video action recognition representation with more semantic language monitoring and achieved zero-shot video action recognition. These outstanding works inspire  exploration of whether and how this alignment is beneficial for ZS-SBIR.
    
\section{METHOD}

    In this section, we first briefly describe the ZS-SBIR problem setting (Sec.~\ref{sec: Problem Setting}). Then, we introduce our main ideas gradually to show how we proposed our model from naive adaption to effective adaption and vision-language alignment (Sec.\ref{sec: Our Method}). Finally, we summarize the objective of our model.

\subsection{Problem Setting}\label{sec: Problem Setting}
    
    In the ZS-SBIR setting, the dataset comprises a training and testing subset. As we have two domains samples, \textit{i.e.}, sketches and photos, we denote $\mathcal{D}_{tr}=\{\mathcal{P}^{seen},\mathcal{S}^{seen}\}$ as the training subset where $\mathcal{P}^{seen}$ and $\mathcal{S}^{seen}$ represent images and sketches for seen classes respectively. Similarly, the testing set is denoted as $\mathcal{D}_{te}=\{\mathcal{P}^{unseen}, \mathcal{S}^{unseen}\}$ which been utilized for validating the retrieval performance. We further define $\mathcal{P}^{seen}=\{(p_i,y_i) | y_i \in \mathcal{C}^{seen}\}_{i=1}^{n_{1}}$ and $\mathcal{S}^{seen}=\{(s_i,y_i) | y_i \in \mathcal{C}^{seen}\}_{i=1}^{n_{2}}$, where $y$ represents the category label and $n_1, n_2$ denote the numbers of photos and sketches respectively. $\mathcal{C}^{seen}$ denotes seen classes set. Mathematically, this definition can also be extended to the unseen subset. Note that under the zero-shot scenario, the $\mathcal{C}^{seen}$ and $\mathcal{C}^{unseen}$ are disjoint.

    During the training phase, the ZS-SBIR model is trained on $\mathcal{D}_{tr}$. After trained, it is expected to retrieve images ${p_j} \subseteq \mathcal{P}^{unseen}$ that have same label with the given query sketch $s_i \in \mathcal{S}^{unseen}$, \textit{i.e.}, $y_j = y_i$.

\subsection{Our Method}\label{sec: Our Method}
    
    \noindent\textbf{Overall Architecture}.
    The key solution for cross-domain retrieval is to generate domain-agnostic representation in a shared space. Given the success of the parameter-efficient fine-tuning method and the feature alignment in the foundation model, in this work, we study how to efficiently balance domain heterogeneity and align vision-language in semantic space. As illustrated in Figure~\ref{fig: framework for our method} (b), we employ a teacher-student architecture recommended by SAKE~\cite{liu2019semantic}. Specifically, the teacher and student are initialized from the same model. We conduct distillation learning on the source domain (\textit{i.e.}, ImageNet-1k~\cite{ImageNet-1k}) and discrimination learning on the target domain\footnote{We adhere to the general concept that the source domain refers to the domain of pre-training datasets, while the target domain pertains to the domain of downstream datasets.}. We noticed that our method has two notable characteristics. Firstly, we propose an improved adaptation strategy by incorporating a few trainable parameters to facilitate learning of domain-consistent features and enhance transferability in downstream tasks. Secondly, we focus on learning high-level semantically-aligned features and consider modeling it as an image-text similarity matching problem following the success of foundation models such as CLIP.
    
    \noindent\textbf{Naively Adapting in Downstream Task}.
    Recently, CLIP-Adapter~\cite{gao2021clip-adapter} demonstrates that incorporating additional trainable parameters at the top of the CLIP image encoder can improve generalization ability. Building on this notion, we introduce a simple yet useful baseline that adds two FC layers $h(\theta_2)$ on top of backbone $g(\theta_1)$.  Our goal is that adapt to the new domain and adjust the feature dimension by $h(\theta_2)$. Then, We project to common feature space by $\mathbf{F}(\Theta_S): \mathbb{R}^{H\times W \times C} \rightarrow \mathbb{R}^d$ where $\mathbf{F} \left( \Theta_S \right)=g(\theta_1) \circ h(\theta_2)$. Given an image $x^i \in \mathbb{R}^{H\times W \times C}$, vision features can be written as: $f^{sk/im}=\mathbf{F}\left( x^i;\Theta_S \right) \text{where} \quad x^i\in \left\{ s^i,p^i \right\} $. We will not modify the teacher in the next step as we only need the output logits about the source domain.
    
    In detail, we have two sample objective functions for discrimination and distillation learning. Firstly, we use a conventional strategy that trains a classifier from scratch. For a given image $x^i\in \left\{ s^i,p^i \right\}$, the objective can be written as 
    \begin{align}
        \sigma\left(z\right) &= \frac{\exp \left(z \cdot w_i + b_i \right)}{\sum_{j=1}^{K} \exp \left(z \cdot w_j + b_j \right)}, \\
        \mathcal{L}_{cls} &= \frac{1}{N} \sum_{i=1}^{N} \mathcal{L}_{CE}\left( \sigma(\mathbf{F}\left(x^i; \Theta_S \right) / \tau); y_i\right), 
    \end{align}
    where the $w$ and $b$ are the weight and bias terms in the benchmark label classifier $\mathbf{W}$. Meanwhile, $y$, $\tau$, $\sigma(\cdot)$ and $\mathcal{L}_{CE}$ denote the ground truth label, the temperature coefficient, \textit{softmax} function and standard Cross-Entropy Loss respectively.  We aim to learn intra-class aggregation and inter-class separation properties.

    Secondly, we expect to maintain the transferable ability and overcome catastrophic forgetting by logit-level distillation learning. We achieve this by the following objective function:
    \begin{equation}
        \begin{split}
            \hat{y}_i &= \sigma\left(g\left(x_i; \Theta_T \right)\right), \\
            \mathcal{L}_{distill} &= \frac{1}{N} \sum_{i=1}^{N} \mathcal{L}_{CE}\left( \sigma(\mathbf{F}\left(x^i; \Theta_S \right)); \hat{y}_i\right).
        \end{split}
    \end{equation}
    Specifically, we use teacher's predictions in pre-train datasets $\hat{y}_i$ to supervise student.
    
    \noindent\textbf{Effective Domain Adapter}.
    Liu \textit{et al}.~\cite{liu2019semantic} believed that preserving the source domain prior knowledge can improve the generalization ability. Complementary to their novelty, we address it by regarding better adaption to downstream tasks. We believe only naively adding two FC layers is insufficient and insert adapters in our model due to its simplicity as` our objective is to investigate the benefits of the parameter-efficient tuning method. As shown in Figure~\ref{fig: framework for our method} (a), an adapter is added after each ResNet stage or ViT layer. It adopts a bottleneck architecture and can be expressed as follows:
    \begin{equation} \label{equation: Adapter}
        Y = X + \left(ReLU \left(X\cdot W_1\right)\right)\cdot W_2,
    \end{equation}
    where $W$ can be $1 \times 1$ convolution or FC layer. We denote our network as $\mathbf{F}(\hat{\Theta}_S)$ after insert adapters. In contrast, Adapter~\cite{adapter} keeps the pre-trained model frozen and only fine-tunes the additional parameters, but we fine-tune the entire network as we demonstrate finetuning entire model is coast-effective.
    
    \noindent\textbf{Vision-Language Alignment}.
    The success of CLIP in many zero-shot and few-shot tasks has demonstrate that semantic alignment is helpful for knowledge transfer. Firstly, we review the method of CLIP.
    
    CLIP contains two separate encoders (\textit{i.e.}, image and text encoder) and aims to extract high-level visual and textual representation. The image encoder $\mathbf{V}$ can be CNN or ViT, while the text encoder $\mathbf{T} $ is Transformer~\cite{transformer}. As for the visual component, an image is first divided into fixed-size patches and tokenized $E=\{v_j\}_{j=0}^{m} ; v_j \in \mathbb{R}^d$. Then a learnable [class] token is prepended as $E=\{E; v_{cls}\} \in \mathbb{R}^{(m+1)\times d}$ and position encoding $\{v_j^{pos}\}_{j=0}^{m+1}; v_j^{pos} \in \mathbb{R}^d$ is applied. These token sequences are passed to the visual encoder to extract image features denoted by $f_i=\mathbf{V}(x_i)$. Similarly, a sentence is also tokenized by parsing to separate tokens, i.e., $E=\{e_j\}_{j=0}^{q}; e_j \in \mathbb{R}^d$ and appended a learnable class token to form input matrix $E=\{E; e_{cls}\} \in \mathbb{R}^{(q+1)\times d}$. After applied position embedding, the entire process can be denoted as $t_i = \mathbf{T}(s_i)$ where $s_i$ is the input sentence. Ultimately, the training objective can be described as follows: 
    \begin{equation} \label{equ:CLIP}
        \mathcal{L}=\frac{1}{N}\sum_{i=1}^N{y}_i\log \frac{\exp \left( \cos \left( f_i\cdot t_j \right) /\tau \right)}{\sum_{j=1}^K{\exp}\left( \cos\left( f_i\cdot t_j \right) /\tau \right)},
    \end{equation}
    where $y_i$ is groundtruth label and $\cos(f\cdot t)$ means cosine similarity and $\tau$ is the temperature coefficient. In the zero-shot reference, CLIP classifies and translates it into logits by assessing the similarities between the image features and the text features. 
    
    The superiority of CLIP's zero-shot ability is attributed to its open vocabulary prompt and explicit feature alignment. Given this, we utilize the CLIP text encoder to extract textual features. Subsequently, in a manner akin to CLIP, we employ a simple dot product and contrastive loss (equation~\ref{equ:CLIP}) to align our student model with the CLIP text encoder.

    Recently, prompt-tuning methods~\cite{jia2022visual, prefix-tuning, Black-Box-Tuning} have gained significant attention due to their better performance and parameter-efficient nature, which sparked our curiosity about can these properties be used for sketch-related tasks. Diverging from CLIP-AT’s approach of incorporating trainable tokens into the CLIP image encoder, we conducted additional investigations into the semantic knowledge inherent in category names by resorting to a more semantic text encoder. We examined prominent textual prompts methods, including learnable prompt (\textit{e.g.}, CoOp) and other prompt engineering approaches. Based on our experimental results, we selected the hand-prompt as it fits our datasets best, \textit{i.e.}, hand designed and fixed templates: \texttt{a photo of [class]}. We first define a template fill function as $Z_{fill}(\mathbf{S}, c)$ to construct different prompts, where $c$ is the category name and $\mathbf{S}$ means prompt templates. Then, we extract the text feature using CLIP text encoder as $t_i= \mathbf{T}(Z_{fill}(\mathbf{S},c_i))$. For visual part, we extract image feature as $f_i = \mathbf{F}\left(x^i; \hat{\Theta_S} \right)$. We set the text feature extraction process offline to improve training efficiency. To model the similarity between image and language, we use the text features as classifier and fix it during training. If not specified, we use the text encoder paired with the ResNet-50 image encoder. Then the alignment loss can be written as follows:
    \begin{align}
            \mathbf{z_i} &= [cos(f_i, t_1), cos(f_i, t_1), \cdots cos(f_i, t_C)], \\
            \mathcal{L}_{align} &=\frac{1}{N} \sum_{i=1}^{N} \mathcal{L}_{CE}\left(\frac{\exp \left(\mathbf{z_i}/\tau \right)}{\sum_{j=1}^{C} \exp \left(\mathbf{z_j} /\tau \right)}; y_i\right).
    \end{align}
    We hope the model learned knowledge from $\mathcal{C}^{seen}$ can generalize to $\mathcal{C}^{unseen}$ like CLIP.

\subsection{Overall Objective}
    Combining the above definitions, we train our model end to end with the following objective function:
    \begin{equation}
        \mathcal{L} = \mathcal{L}_{align} + \lambda \mathcal{L}_{distill},
    \end{equation}
    where $\lambda$ is the hyper-parameter. 

    \begin{table*}[ht]
      \caption{Comparison with state-of-the-art. The symbol $^{\dag}$ and $^{\ddag}$ note ResNet and DINO-based model respectively. Semantic and dim mean whether to use additional semantic vectors and the feature dimensions. Pretrain means the datasets used to pre-train the model. We highlight our model with \colorbox{sherrytan!50}{tan}. The best result is in \textbf{bold}, second-best result is \underline{underlined}}
      \label{tab: Compare With sota}
      \centering
      \setlength{\tabcolsep}{2.5mm}{
          \begin{tabular}{p{35mm}<{\centering} p{8mm}<{\centering} p{5mm}<{\centering}ccccccc}
            \toprule[1.5pt]
            \multirow{2}*{Methods} & \multirow{2}*{Semantic} & \multirow{2}*{Dim} & \multirow{2}*{Pretrain} & \multicolumn{2}{c}{Sketchy Ext split1} & \multicolumn{2}{c}{Sketchy Ext split2} & \multicolumn{2}{c}{TU-Berlin} \\ 
            & & & & mAP@all & Prec@100 & mAP@200 & Prec@200 & mAP@all & Prec@100 \\
            \midrule[1pt]
            \demphs{SAKE (ICCV-19)~\cite{liu2019semantic}} & \demphs{\cmark} & \demphs{512} & \demphs{IN-1K} &\demphs{54.7} & \demphs{69.2} & \demphs{49.7} & \demphs{59.8} & \demphs{47.5} & \demphs{59.9}\\
            
            DSN (IJCAI-21)~\cite{wang2021domain}& \cmark & 512 & IN-1K &58.3 & 70.4 & \textbf{--} & \textbf{--} & 48.1 & 58.6\\
            TCN (TPAMI-21)~\cite{wang2021transferable} & \cmark & 512 & IN-1k &61.6 & \textbf{76.3} & \underline{51.6} & \underline{60.8} & 49.5 & \underline{61.6}\\
            StyleGuide (TMM-21)~\cite{dutta2020styleguide}& \cmark & 200 & IN-1K & 37.6 & 48.4 & \textbf{--} & \textbf{--} & 25.4 & 35.5\\
            RPKD (ACM MM-21)~\cite{tian2021relationship} & \xmark & 512 & IN-1K &61.3 & 72.3 & 50.2 & 59.8 & 48.6 & 61.2\\
            NAVE (IJCAI-21)~\cite{wang2021norm} & \xmark & 512 & IN-1K &61.3 & 72.5 & \textbf{--} & \textbf{--} & 49.3 & 60.7\\
            $\text{PSKD}^{\dag}$ (ACM MM-22)~\cite{wang2022prototype} & \xmark & 512 & IN-1K+ & \underline{62.7} & 75.0 & 48.6 & 58.2 & 41.9 & 60.8\\
            \rowcolor{sherrytan!50}$\text{Sherry}^{\dag}$ & \cmark & 512 & IN-1K & \textbf{65.5} & \underline{75.3} & \textbf{54.0} & \textbf{63.2} & \textbf{52.2} & \textbf{62.6}\\
            \hline
            TVT (AAAI-22)~\cite{tian2022tvt} & \xmark & 384 & IN-1K+ & 64.8 & 79.6 & 53.1 & 61.8 & 48.4 & {66.2}\\
            $\text{PSKD}^{\ddag}$ (ACM MM-22)~\cite{wang2022prototype} & \xmark & 384 & IN-1K+ & 68.8 & 78.6 & 56.0 & 64.5 & 50.2 & {66.2}\\
            \lemphs{CLIP-AT (CVPR-23)~\cite{sain2023clip}} & \lemphs{\cmark} & \lemphs{768} & \lemphs{WIT} &\lemphs{--} & \lemphs{--} & \textbf{\lemphs{72.3}} & \textbf{\lemphs{72.5}} & \textbf{\lemphs{65.1}} & \textbf{\lemphs{73.2}} \\
            \lemphs{ZSE (CVPR-23)~\cite{ZSE-SBIR}} & \lemphs{\xmark} & \lemphs{768} & \lemphs{IN-1K} &\lemphs{69.8} & \lemphs{79.7} & \lemphs{52.5} & \lemphs{62.4} & \lemphs{\underline{54.2}} & \lemphs{65.7} \\
            \rowcolor{sherrytan!50}$\text{Sherry}^{\ddag}$ & \cmark & 384 & IN-1K & \textbf{74.1} & \textbf{83.5} & \underline{61.6} & \underline{69.5} & 54.1 & \underline{66.4}\\
          \bottomrule[1.5pt]
        \end{tabular}
    }
    \end{table*}
    
    
\section{EXPERIMENTS}
\subsection{Data and Setting}
    \textbf{Dataset}. Fallowing the existing art~\cite{SounakDey2019DoodleTS, wang2022prototype, tian2021relationship}, we evolutate our method in three popular benchmark datasets including Sketchy~\cite{sangkloy2016sketchy}, TU-Berlin~\cite{tuberlin} and QuickDraw~\cite{SounakDey2019DoodleTS}.

    \textit{Sketchy} is a large-scale dataset of fine-grained aligned sketch-image pairs. We use its extended version~\cite{LiLiu2017DeepSH} that contains 75,471 sketches and 73,002 nature images in 125 categories. For a fair comparison, we follow the split method proposed in~\cite{YumingShen2018ZeroShotSH} and~\cite{ZS-SBIR-framework}, randomly selecting 25/21 and 100/104 categories as the testing set and training set. To be easily distinguished, We denote split1 and split2, respectively.  

    \textit{TU-Berlin} contains 20,000 sketches over 250 categories and 13,419 natural images, additional 191,067 nature images collected by Zhang \textit{et al.}~\cite{zhang2016sketchnet} and finally yielding a total of 204,489 photos. It is a highly imbalanced dataset as the number of sketches is only one-tenth that of images. Meanwhile, it has a higher level of abstraction for sketches. Following ~\cite{YumingShen2018ZeroShotSH}, we select 30 categories for testing and another 220 for training.

    \textit{QuickDraw} is a new large-scale dataset, which is a huge collection of drawings belonging to 345 categories collected from \emph{Quick, Draw!}\footnote{\href{https://quickdraw.withgoogle.com/}{https://quickdraw.withgoogle.com/}} game. As sketches are produced in an amateur drawing style, it has an extensive domain gap between non-expert drawers and raw photos. Dey~\textit{et al.}~\cite{SounakDey2019DoodleTS} selected 110 categories containing 330,000 sketches and 204,000 images and separated 30/80 categories for testing and training. 

    \noindent\textbf{Implementation Details}. We implemented our method in the PyTorch toolkit with one RTX3090 GPU. In the experiment, we use CSE-ResNet-50 and DINO-s/8 as our backbone. If not specified, we use ResNet and DINO to indicate different models with different backbones. We use the pre-trained model on ImageNet-1k provided by~\cite{caron2021DINO} and SAKE~\cite{liu2019semantic} to initialize our model. For both backbones, we set the maximum training epochs as 40 and use the Adam optimizer with the weight decay of 5e-4. We use the same augmentation strategy as SAKE~\cite{liu2019semantic}. 

    \noindent\textbf{Evaluation Protocol}. During inference, we only use student model to extract the image feature and do not include any text embedding for fair comparison. We evaluate our model by adopting the same evaluation protocol as previous art~\cite{liu2019semantic, tian2022tvt}, \textit{e.g.}, mean average precision (mAP@k) and precision (Prec@k). We use cosine similarity between sketches and photos as the distance metric to compute the retrieval results as SAKE~\cite{liu2019semantic}. 
    \begin{figure}[t]
        \centering
        \includegraphics[width= 0.48\textwidth]{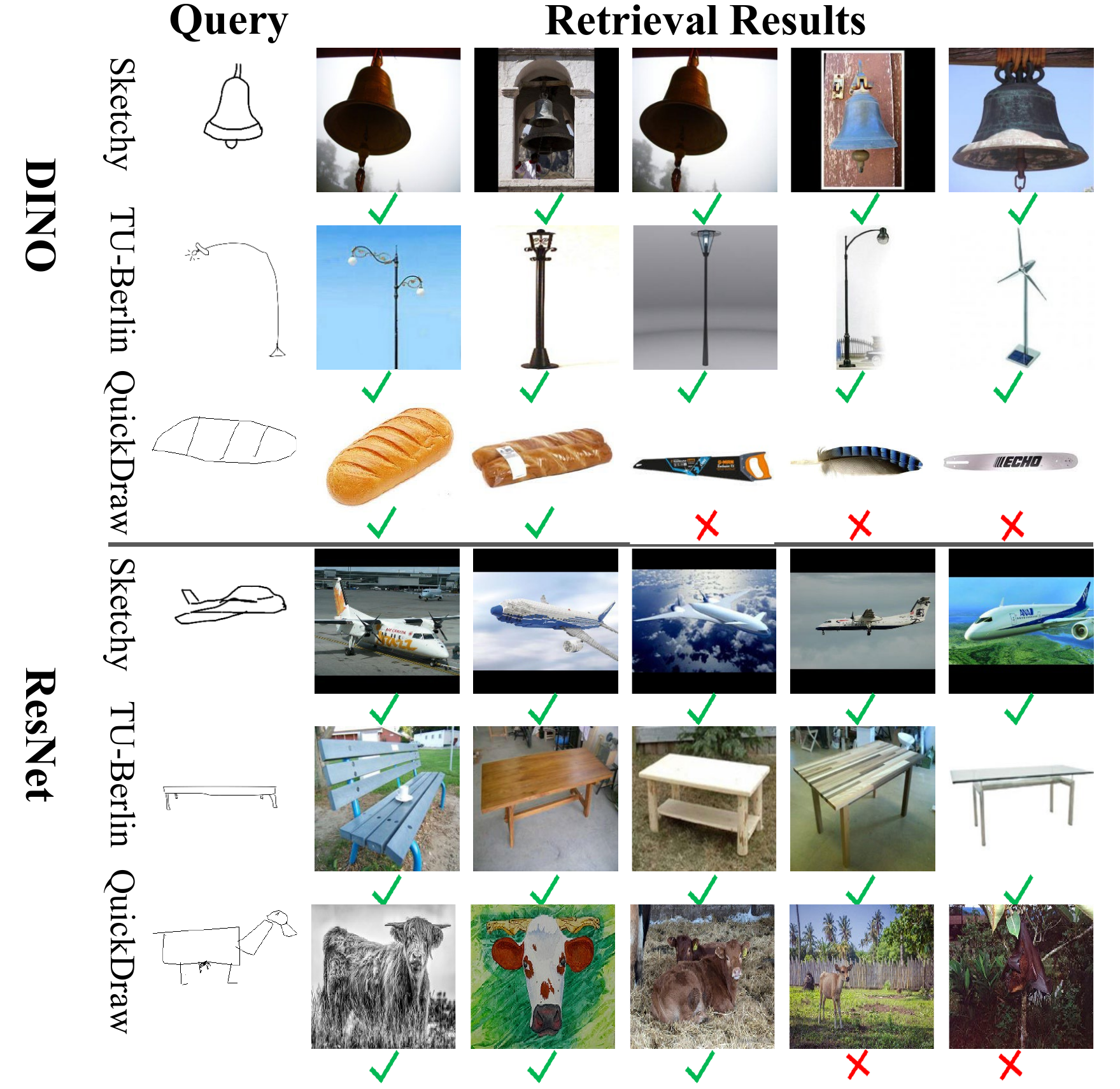}
        \caption{Retrieval result on three datasets. Sherry can steadily and correctly retrieve the unseen categories on Sketchy and TU-Berlin but retrieve not well in QuickDraw as it is the most serious difficulty for it. Ranked by similarity, we selected 5 retrieved photos for each random sketch query. All the results are marked at the bottom respectively.}
        \label{fig: detail retrieval result}
    \end{figure}
    
\subsection{Comparison with State-of-the-Arts}

    We compare two types of models on the benchmark datasets, \textit{i.e.}, models with and without semantic vectors. Notably, except for our method, almost all the models incorporating semantic information use hierarchical mode~\cite{miller1995wordnet} or Word2vec~\cite{mikolov2013word2vec}. However, we utilize the dense text features to achieve alignment in CLIP feature space. The overall comparison result can be seen in Table~\ref{tab: Compare With sota} and Table~\ref{tab: Compare on QucikDraw Dataset}. We interpret IN-1K+ means pre-trained on ImageNet-1k and downstream sketch benchmarks. \lemphs{CLIP-AT} use CLIP image encoder that contains the prior WIT knowledge and is likely to cause imprecise zero-shot setting. Both \lemphs{CLIP-AT} and \lemphs{ZSE} use a larger ViT-B backbone (87 million parameters larger than our 23 million parameters). So, direct comparison against our approach is not feasible as potential bias arise from the larger dataset (400 million) or model capacity. However, we also achieve better results in a reasonable way.
    
    We compare Sherry and previous works in Table~\ref{tab: Compare With sota} and divide the results with a horizontal line according to different backbones. The ResNet-based models are displayed above it, while ViT-based models are below. We achieve new state-of-the-art results in two different backbones regardless of the difficulty of the datasets, demonstrating that our adapter and alignment strategy enjoys effectiveness and applicability to various frameworks.
    
    For the  models that do not incorporate semantic information, we believe the reason for their slightly worse performance is a lack of alignment between visual and semantic features. In other words, during training, images are simply mapped as one-hot vectors, or only the distance in one modality (\textit{e.g.}, image) is tightened. Furthermore, compared with other models that incorporate semantic information, our vision-language-aligned model surpasses these significantly. This means the jointly learned text embeddings of visual and language data can better encode visual similarities than those learned from linguistic corpora alone (\textit{e.g.}, WordNet, Word2vec). Compared with methods that fine-tuned teacher model before training to accommodate the abstract nature of sketches, our method avoids this expensive operation and achieves better performance thanks to our adaption strategy. We only add a few learnable parameters and use simple dot-product operation, making our method much more concise than previous arts.

    \begin{table}[ht]
      \caption{Comparison with state-of-the-arts on QucikDraw. We show our best results using the DINO backbone. The best result is in bold, second best result is \underline{underlined}}
      \label{tab: Compare on QucikDraw Dataset}
      \centering
      \setlength{\tabcolsep}{1.6mm}{
          \begin{tabular}{ccccc}
            \toprule[1pt]
            Method & mAP@all & mAP@200 & Prec@100 & Prec@200\\
            \midrule[0.75pt]
            \demphs{doodle~\cite{SounakDey2019DoodleTS}} & \demphs{7.5} & \demphs{9.0} & \demphs{--} & \demphs{6.8} \\
            RPKD~\cite{tian2021relationship} & 14.3 & 12.8 & 23.0 & 21.8 \\
            TVT~\cite{tian2022tvt} & 14.9 & 19.1 & \underline{}{29.9} & 29.3 \\
            PSKD~\cite{wang2022prototype} & 15.0 & \textbf{19.9} & 29.7 & \underline{29.8} \\
            \lemphs{ZSE~\cite{ZSE-SBIR}} & \lemphs{14.5} & \lemphs{--} & \lemphs{--} & \lemphs{21.6}\\
            \lemphs{CLIP-AT~\cite{sain2023clip}} & \textbf{\lemphs{20.2}} & \lemphs{--} & \lemphs{--} & \textbf{\lemphs{38.8}}\\
            \rowcolor{sherrytan!50}$\text{Sherry}^{\ddag}$ & \underline{18.0} & \underline{19.5} & \textbf{31.3} & \underline{29.8} \\
          \bottomrule[1pt]
        \end{tabular}
    }
    \end{table}

    \noindent\textbf{Retrieval Examples}.
    We calculate the cosine similarity to rank the candidates and select the final retrieval results.
    
    In Figure~\ref{fig: detail retrieval result}, We show our retrieval results on three datasets. Sherry can steadily achieve high-quality retrieval in two common scenes: the Sketchy and TU-Berlin datasets, but poorly retrieved in QuickDraw. Requiring the model to differentiate fine-grained information and be equipped with higher robustness as QuickDraw consists of large amounts of raw and noisy data (e.g., misaligned resolution, small objects, high-level noisy background, abstract concepts). Thus, we think exploring how to deal with noisy data and learn fine-grained semantics is a solution, but we didn't attempt that in this work.



\begin{figure}[ht]
        \centering
        \includegraphics[width=0.45 \textwidth]{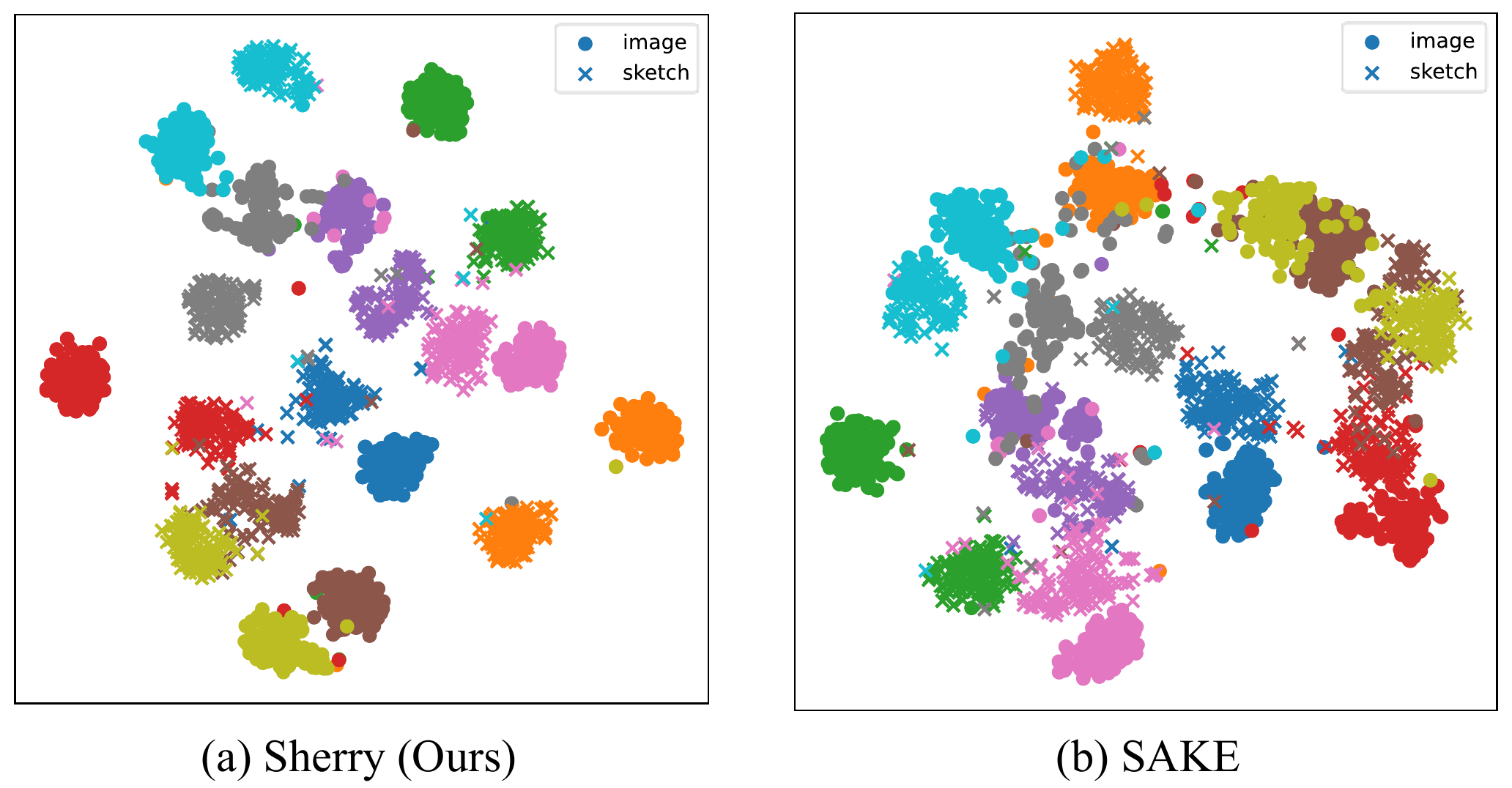}
        \caption{The t-SNE visualization of features. Comparison under same CSE-ResNet-50 backbone and use same random categories. The results show that we make a more compact intra-class and more discriminative inter-class feature distribution. Each color represents a category respectively.}
        \label{fig: t-SNE result}
    \end{figure}

    \noindent\textbf{Visualization of Features}.
    To demonstrate the excellent and compact characteristics of our model in feature space, we choose the t-SNE~\cite{van2008visualizing} algorithm to visualize our retrieval feature. We compare our method with SAKE by randomly selecting ten unseen categories and 100 sketches and photos for each category. As seen in Figure~\ref{fig: t-SNE result}, SAKE lacks regularity in feature space, and the distribution of unseen classes is not well-converged. In contrast, our model performs well, with similar photos and sketches converging closely. We argue that a well-clustered feature space means a better retrieval performance. This result indicates that our model has a more compact intra-class and separated inter-class feature distribution. We attribute this phenomenon to our adapter strategy has balanced the domain bias between sketches and photos.

\subsection{Analysis for Model}
    We further analyze our domain adapter and vision-language alignment strategy in this section. 

    \noindent\textbf{Why Adapter}: Compared to some models that apply the Adapter, our backbone with 23 million parameters is small. For example, VPT~\cite{jia2022visual} and AIM~\cite{AIM} use ViT-B/16 or even the larger foundation models, such as ViT-L/16 for downstream tasks, with 87 and 304 million parameters, respectively. As our goal is to achieve better accuracy in cross-domain retrieval, it is affordable and cost-effective to fine-tune the entire model.
    Although our model is smaller in size, adapter can also bring inspiring benefits in our ablation experiment as those larger models in many scenarios. First, as shown in Table~\ref{tab: Analysis For Adapter}, freezing backbone and inserting a few adapters that only 0.8 million parameters are trainable (Head+Adapter) under the DINO-based model, we achieve higher mAP@all compared with PSKD ($68.95\%$ vs. $68.8\% $).  Compared to the model that only the "Head" is tunable, our ResNet-based model achieves improvements of \textcolor{green!80}{$31.1\%$} and \textcolor{green!80}{$30.3\%$} on two datasets by adding few adapters (Head vs. Head+Adapter). These improvements can be even higher when using the DINO-based model. It indicates that adapters can effectively learn to transfer knowledge to new downstream tasks while preserving prior knowledge. However, it still has a large gap with the full fine-tuned model. It can be seen that a full fine-tuned model (Backbone and Backbone+Adapter) can bring significant improvement compared to those only a few parameters are tunable. We attribute this to the prior domain bias problem of frozen prior knowledge and lacking new abstract concepts for sketches. In other words, it only focuses on samples from dominant domains (\textit{i.e.}, photos). Finally, our full fine-tuned model aims to overcome the problem above. Our ResNet-based model achieves impressive improvements of \textcolor{green!80}{$2.9\%$} and \textcolor{green!80}{$2.5\%$} in two different datasets by adding and training adapters (Backbone vs. Backbone+Adapter). These results successfully validate the effectiveness of the proposed adaptation strategies.
    
    \begin{table}[ht]
      \caption{Effectiveness of proposed adapter. We have two settings for adding adapter ablation in two different backbones: fine-tuning a few projection layers or the entire model. Param means the number of parameters in millions. "Head" means a few projection FCs layers on top of the backbone. "+Adapter" means insert adapters in framework.}
      \label{tab: Analysis For Adapter}
      \centering
      \setlength{\tabcolsep}{1mm}{
          \begin{tabular}{cccccc}
            \toprule[1pt]
            Model & Tunable & Param & \makecell{Tunable \\ Param} & \makecell{Sketchy \\ split1} & TU-Berlin \\
            \midrule[0.75pt]
            \multirow{4}*{ResNet} & Head & 29.5 & 3.3 & 22.52 & 11.22\\
              & Head+Adapter& 32.3 & 6.0 & 53.64\dtplus{+31.1} & 41.55\dtplus{+30.3}\\
              \cline{2-6}
              & Backbone & 29.5 & 29.5 & 62.55 & 49.77\\
              & Backbone+Adapter & 32.3 & 32.3 & 65.49\dtplus{+2.9} & 52.24\dtplus{+2.5}\\
            \hline
            \multirow{4}*{DINO} & Head & 24.5 & 2.8 & 12.17 & 7.68\\
              & Head+Adapter& 25.3 & 3.6 & 68.95\dtplus{+56.78} & 41.46\dtplus{+33.8}\\
              \cline{2-6}
              & Backbone & 24.5 & 24.5 & 72.05 & 52.58\\
              & Backbone+Adapter & 25.3 & 25.3 & 74.1\dtplus{+2.1} & 54.1\dtplus{+1.5}\\
          \bottomrule[1pt]
        \end{tabular}
    }
    \end{table}
    
\begin{figure}[ht]
        \centering
        \includegraphics[width=0.45 \textwidth]{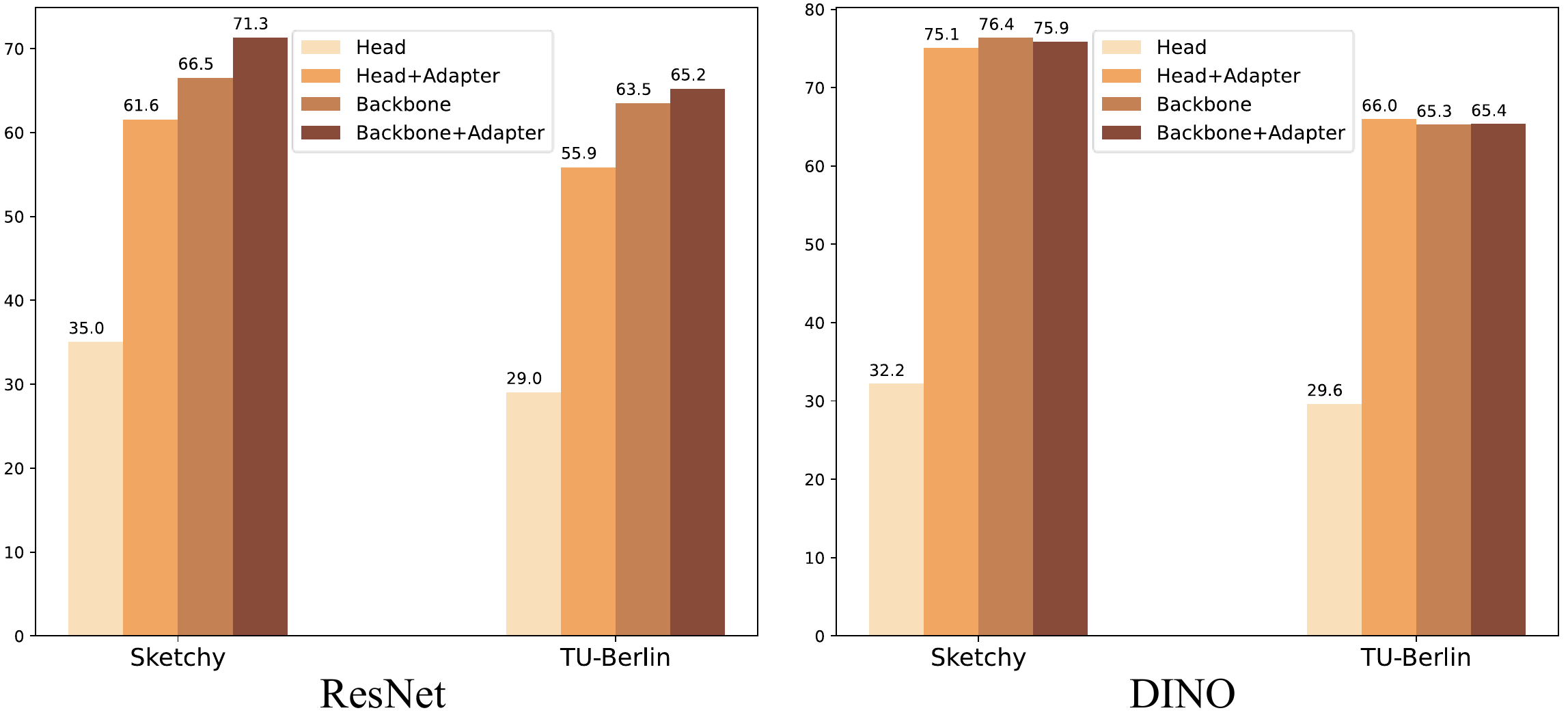}
        \caption{The ZS-SBSR mAP@all result on Sketchy and TU-Berlin. A steady boost can be seen after insert adapters for two models. This improvement shows that the adapter can help the model learn more about the sketch domain, so it can moderate domain bias in SBIR task}
        \label{fig: SBSR result}
    \end{figure}

    To further showcase the efficiency of our adapter in facilitating comprehension of new concepts about sketches, we explore zero-shot sketch-based sketch retrieval (ZS-SBSR). In other words, given a sketch query, we aim to retrieve sketches from the same class. We randomly select some sketches as queries for each unseen category. Then, the rest of the sketches will be used as a search gallery to ensure no overlap with the query. We use mAP@all to evaluate the ZS-SBSR results. In this case, the model capable of recognizing abstract concepts depicted in sketches will result in favorable ZS-SBSR results. Figure~\ref{fig: SBSR result} demonstrates that adding adapter and training together with head can significantly improve in two datasets (Head vs. Head+Adapter). Meanwhile, the result is slightly lower than the full fine-tuned model for the ResNet-based case but comparable to the DINO-based case (Head+Adapter vs. Backbone). It means lightweight adapter can learn more sketch information. In the full finetuning case, our ResNet-based model still achieves a steady boost (66.5 vs. 71.3). This indicates that combining the backbone with adapters can effectively accommodate the new abstract domain and learn the balance between images and sketches.  

    \noindent\textbf{Why Vision-Language Alignment}. We conducted experiments on text template selection to analyze the impact of the vision-language alignment strategy.
    
    It has been demonstrated in CLIP that vision-language alignment is beneficial for alleviating the semantic gap between seen and unseen classes. We believe this also works well in ZS-SBIR task. We choose four different kinds of prompt methods that can be seen in Table~\ref{tab: Analysis For text prompt}. For the prompt ensembling, it is a subset of ViLD~\cite{gu2021ViLD}, and we made some modifications to adapt the sketch samples. CoOp-16 means the learnable pseudo words length is 16 (CoOp introduced learnable textual contexts to achieve better transferability by directly optimizing the contexts using back-propagation. Details can be seen in~\cite{zhou2022CoOp}). We observe a robust improvement in both datasets using any of these prompt methods. It means our alignment with dense text bring a more semantic feature space. Meanwhile, the potential space of CLIP helps us transfer knowledge from seen classes to unseen classes since we align with the open-vocabulary CLIP text encoder. Thus, our method is better than other models for zero-shot ability. Our experiments concluded that the more naive prompt methods (first and second row) bring more significant performance improvements. We attribute this phenomenon to the samples of these two datasets, which mainly contain single objects and no complex backgrounds (seen in Figure~\ref{fig: detail retrieval result}). It means a prompt that is more consistent with the dataset prior can provide incredible benefits to the multi-modality model during the alignment process.
    
    \begin{table}[ht]
      \caption{Analysis of textual prompts on Sketchy and TU-Berlin. We explore ablations on 5 kinds of prompt methods for vision-language alignment. Classical means randomly initial classifier. We use mAP@all as metric to compare all prompt methods.}
      \label{tab: Analysis For text prompt}
      \centering
      \setlength{\tabcolsep}{1.6mm}{
          \begin{tabular}{cccc}
            \toprule[1pt]
            Model & Method & Sketchy split1  & TU-Berlin\\
            \midrule[0.75pt]
            \multirow{5}*{ResNet} & classical & 60.92 &  50.44 \\
            & \texttt{a [class]} & 65.19\dtplus{+4.3} & 51.4\dtplus{+1.0} \\
            & \texttt{a photo of [class]}  & 65.49\dtplus{4.6} & 52.24\dtplus{1.8}  \\
            & prompt ensembling & 64.19\dtplus{+3.3} & 51.25\dtplus{0.8}  \\
            & CoOp-16~\cite{zhou2022CoOp} & 64.21\dtplus{+3.3} & 51.26\dtplus{+0.8}  \\
            \hline
            \multirow{5}*{DINO} & classical & 60.0 &  47.3 \\
            & \texttt{a [class]} & 73.21\dtplus{+13.2} & 52.7\dtplus{+5.4} \\
            & \texttt{a photo of [class]}  & 74.1\dtplus{+14.1} & 54.08\dtplus{+6.8}  \\
            & prompt ensembling & 73.12\dtplus{+13.1} & 53.04\dtplus{+5.7}  \\
            & CoOp-16~\cite{zhou2022CoOp} & 74.14\dtplus{+14.14} & 53.31\dtplus{+6.0}  \\
          \bottomrule[1pt]
        \end{tabular}
    }
    \end{table}
    
\begin{figure}[ht]
        \centering
        \includegraphics[width=0.5 \textwidth]{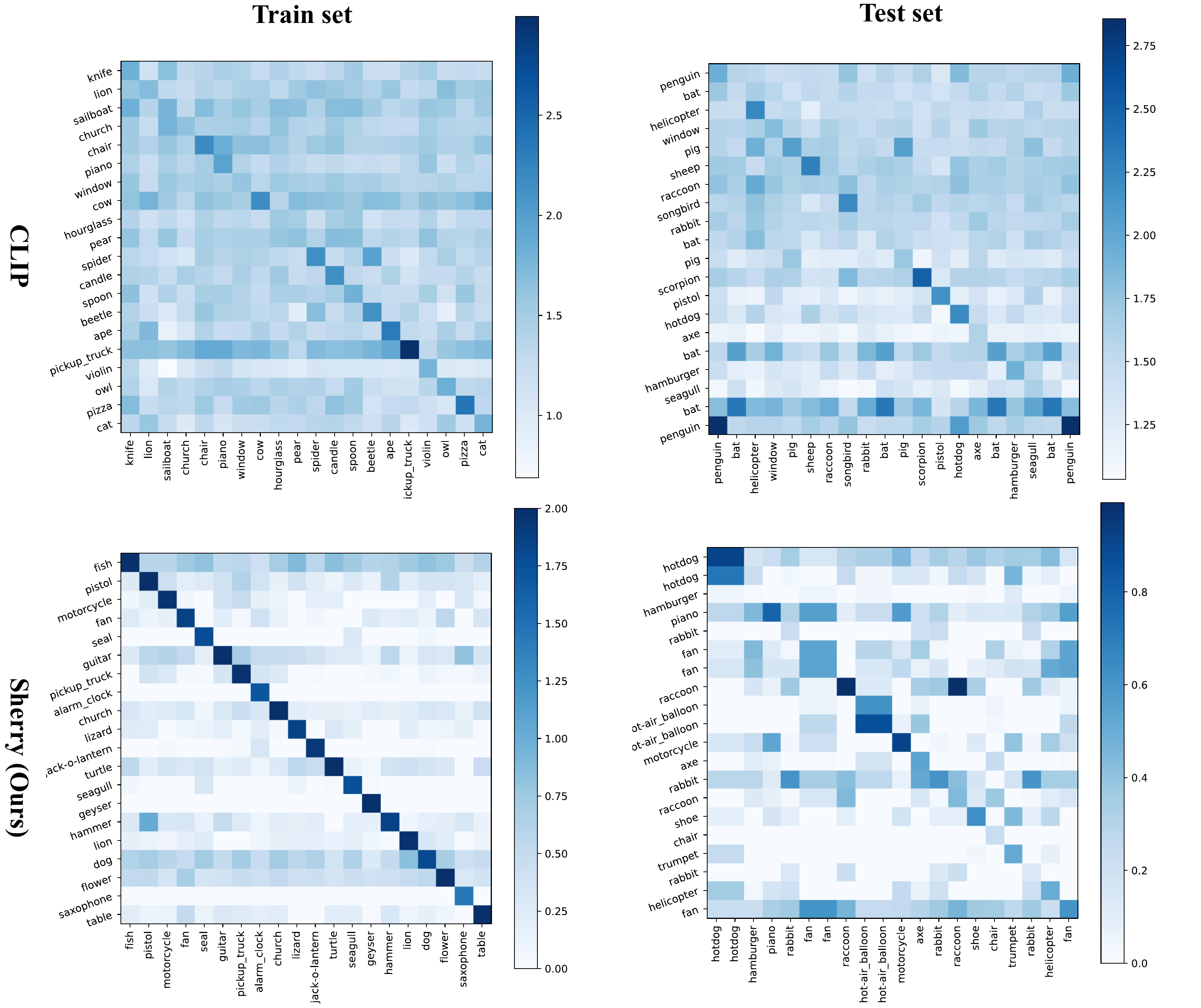}
        \caption{Comparison of vision-language similarity with vanilla CLIP on Sketchy dataset. Diagonal patterns represent paired image-text similarity characteristics. We believe the clearer it is, the better the generalization of the model will be. We randomly select 10 samples for sketches and photos.}
        \label{fig: heatmap_1 result}
    \end{figure}

      As shown in Figure~\ref{fig: heatmap_1 result}, we present a heatmap comparison for the visual-text similarity of the seen and unseen classes between vanilla CLIP and our model. The vertical axis represents the sketches and photos for each heatmap, while the horizontal axis represents the corresponding categories. The top half of each map shows sketch-text similarity, and the bottom half represents photo-text similarity. Note that a more sensible diagonal pattern represents a more precise semantic alignment, which is the critical element for generalization ability. Although vanilla CLIP perform better in other vision tasks, it does not distinguish well between different classes for both seen and unseen categories in the sketch domain. In contrast, our model shows a clear alignment pattern for both seen and unseen classes, indicating that the semantic alignment learned from seen categories has generalized to unseen categories thanks to our straightforward alignment strategy.
     
\subsection{Ablation Study}
    \noindent\textbf{Text Encoder.} Sherry supports arbitrary text encoders in principle. Firstly, we ablate various text encoders that the CLIP provides. Note that all text encoders feature adopt the same transformer-based architecture in CLIP. The main difference between the encoders is the image encoder that was paired during CLIP pre-training.

    We observe that RN50 paired text encoder performs best among all text encoders. We conjecture that this is because our backbone is comparable to CLIP RN50 image encoder in the number of parameters. So, achieving model parameter matching, wherein the text and image encoders possess identical parameters, potentially results in enhanced modality alignment properties when needing the help of the text encoder. 
    
    \begin{table}[ht]
      \caption{Analysis of text encoders on Sketchy and TU-Berlin. We explore the effect of four representative CLIP text encoders on semantic alignment and find RN50 is the best choice for our task. The best result is in \textbf{bold}.}
      \label{tab: Analysis For text encoder}
      \centering
      \setlength{\tabcolsep}{1.6mm}{
          \begin{tabular}{ccccc}
            \toprule[1pt]
            Backbone & Text encoder & Dim & Sketchy split1  & TU-Berlin\\
            \midrule[0.75pt]
            \multirow{4}*{ResNet} & RN50 & 1024 & \textbf{65.49} &  \textbf{52.24} \\
            & RN50$\times$4 & 640 &  {64.26} & {50.74} \\
            & ViT-B/16  & 512 & 64.27 & 50.28  \\
            & ViT-L/14 & 768 & 65.0 & 52.01  \\
            \hline
            \multirow{4}*{DINO} & RN50 & 1024 & \textbf{74.05} &  \textbf{54.08} \\
            & RN50$\times$4 & 640 &  73.35 & 51.87 \\
            & ViT-B/16  & 768 & 70.19 & 52.47 \\
            & ViT-L/14 & 768 & 72.63 & 52.66  \\
          \bottomrule[1pt]
        \end{tabular}
    }
    \end{table}
    
\begin{figure}[ht]
        \centering
        \includegraphics[width=0.45 \textwidth]{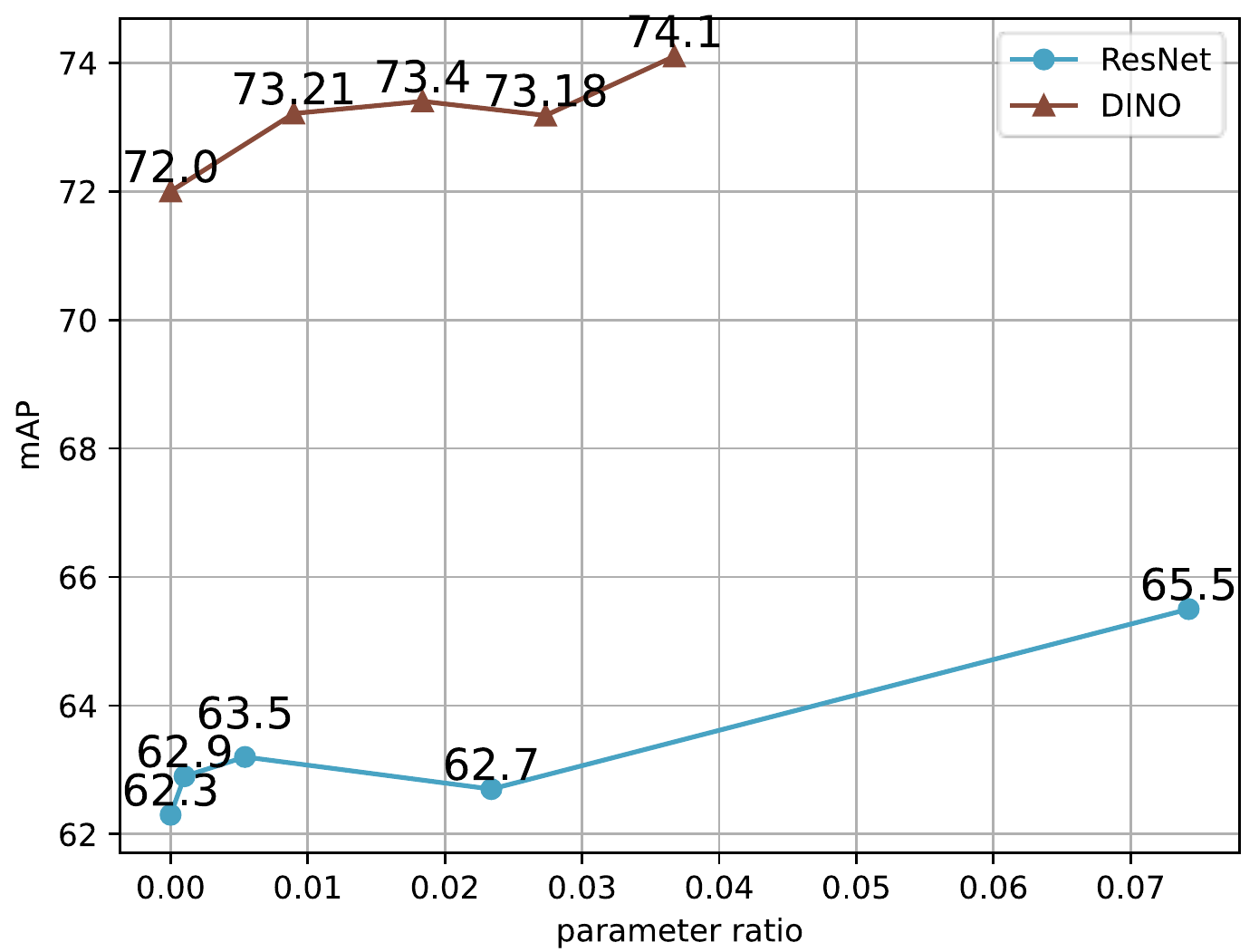}
        \caption{Adapter scalablity analysis. We explore the behaviour of Adapter at scale and find that Adapter is also a good scalable learner to some extent. The horizontal axis represents the ratio of the adapter to the backbone in parameters, and the vertical axis represents the mAP on the Sketchy dataset.}
        \label{fig: adapter_scale}
    \end{figure}

    \noindent\textbf{Scalability of Adapters}. By default, we add Adapter to every ViT block and CSE-ResNet stage (12 blocks and 4 stages in total). To assess its scalability, we gradually increase the number of adapters, starting from zero and progressing toward the default adapter configuration. Considering the ease of implementation, we add the adapter stage by stage for the ResNet-based model. For the DINO-based model, we add the adapter layer by layer. Figure~\ref{fig: adapter_scale} shows results in the Sketchy dataset. We can see that gradually increasing the number of adapters can almost steadily improve the retrieval accuracy for different models. Consequently, we can generalize this observation and assert that, to a certain extent, Adapter is a good scalable learner complementing its inherent parameter-efficiency and lightweight characteristics. This timely realization sheds light on the potential utilization of adapters across numerous downstream tasks.
    
    \noindent\textbf{Two Different Components.} We finally analyze the effect of our adapt and align method using the control variable setting in Table~\ref{tab: Ablation}. We did not study the effects of distillation learning as the evidence presented by SAKE~\cite{liu2019semantic} highlights its indispensability. We observed that each of our singular strategies brings significant improvement, indicating the effectiveness of our strategy in addressing two different challenges in ZS-SBIR from different perspectives. Our baseline model performed poorly since it lacks both adapters and semantic alignment, which is the typical learning pipeline of most existing models. However, we observed a steady improvement when gradually adding our two strategies. It means that our two ideas are complementary and effective.
    \begin{table}[ht]
      \caption{We evaluate mAP@all on Sketchy and TU-Berlin. The symbol "\cmark" indicate this component is used during training, while "\xmark" does not. The results show that our strategy can achieve a stable additive improvement}
      \label{tab: Ablation}
      \centering
      \setlength{\tabcolsep}{1mm}{
          \begin{tabular}{cccccc}
            \toprule[1pt]
             Model & Baseline & Adapter & Alignment & \makecell{Sketchy \\ split} & TU-Berlin\\
            \midrule[0.75pt]
            \multirow{4}*{ResNet}& \cmark & \xmark & \xmark  & 56.49 & 45.46 \\
            & \cmark & \cmark & \xmark  & 60.92\dtplus{+4.4} & 50.44\dtplus{+5} \\
            & \cmark & \xmark & \cmark & 63.56\dtplus{+7.1} & 49.95\dtplus{+4.5} \\
            & \cmark & \cmark & \cmark & 65.49\dtplus{+9.0} & 52.24\dtplus{+6.8} \\
            \hline
            \multirow{4}*{DINO}& \cmark & \xmark & \xmark  & 57.78 & 47.21 \\
            & \cmark & \cmark & \xmark  & 60.0\dtplus{+2.2} & 47.3\dtplus{+0.1} \\
            & \cmark & \xmark & \cmark & 72.05\dtplus{+14.3} & 52.58\dtplus{+5.4} \\
            & \cmark & \cmark & \cmark & 74.05\dtplus{+16.3} & 54.08\dtplus{6.9} \\
          \bottomrule[1pt]
        \end{tabular}
    }
    \end{table}
    
\section{CONCLUSION}
     In this work, we propose Sherry, a simple yet effective method that uses some adapters and a vision-language alignment strategy to address challenges in the ZS-SBIR field. The domain adapter module is parameter-friendly, adding only approximately 1 to 3 million parameters (less than $8\%$ of the original model). Despite its simplicity, it could effectively learn the balance between two domains and significantly improve performance in ZS-SBIR. Additionally, the vision-language alignment strategy can effectively generalize the semantic alignment pattern learned from seen to unseen classes, which is a crucial element in zero-shot learning tasks. Our approach is simple and generally applicable, which can be used for many different frameworks and may benefit from a more powerful backbone or foundation semantic-rich model in the future.
    
    Despite many benefits, there are also limitations in the extremely high vision similarity sense such as QuickDraw. Finally, we hope our work inspires more effective adaptation and semantic transfer strategies.



\bibliographystyle{IEEEtran}
\bibliography{./ref.bib}

\end{document}